%% file: main.tex
\documentclass[10pt,letterpaper,twocolumn]{style}
\usepackage{amsmath,amssymb}
\usepackage{algorithm,algorithmic}
\usepackage{listings}
\usepackage{adjustbox}
\input{authors}

\title{Delving into Asymmetric Information Dynamics for High-Fidelity Virtual Try-On}
\abstract{Virtual try-on (VTON) requires precise pixel-level fidelity, yet mainstream Diffusion Transformers (DiTs) often suffer from texture degradation and structural drift. We identify symmetric interactions in standard joint-attention mechanisms as a source of these failures. Although such interactions support semantic flexibility in general-purpose editing, they allow stochastic noise to corrupt deterministic garment features in VTON. We analyze this problem through asymmetric information dynamics and introduce two diagnostic indicators: Conditional Attention Entropy (CAE) for feature unbiasedness and Injected Information Flux (IIF) for injection effectiveness. Our analysis suggests that symmetric bidirectional attention can corrupt conditional features and attenuate the conditional signal. To address these limitations, we propose \textbf{RealFit}, a framework that combines Unidirectional Information Flow (UIF) with Decoupled Timestep Modulation (DTM). UIF isolates the garment condition from stochastic noise to preserve garment identity, while DTM optimizes the modulation scale to maintain a strong conditional signal. The resulting time-invariant condition branch enables a conditional KV cache that reduces inference time by approximately 75\%. RealFit offers a principled approach to conditional generation with state-of-the-art fidelity and efficiency.}

\begin{document}

\maketitle

\begin{figure*}[t]
\centering
\includegraphics[width=1.0\textwidth]{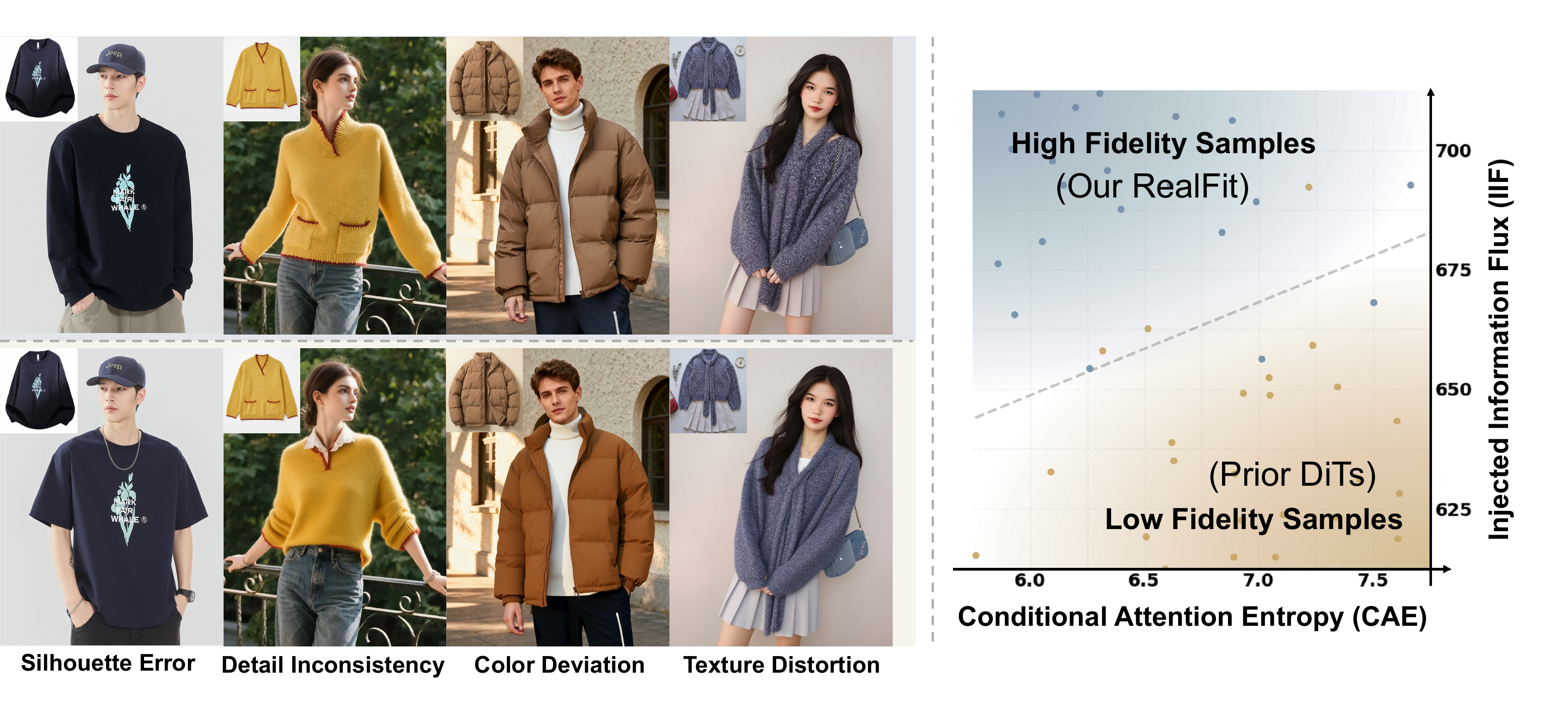}
  \caption{RealFit achieves high-fidelity virtual try-on. (Left) RealFit preserves garment identity across categories, while prior DiTs exhibit errors in silhouette, detail, color, and texture. (Right) Fidelity correlates strongly with CAE and IIF. RealFit reduces CAE and increases IIF to overcome fidelity bottlenecks in symmetric architectures.}
  \label{fig: intro}
\end{figure*}

\section{Introduction}

Image-based virtual try-on (VTON) is an important task in digital fashion and e-commerce. Unlike general image editing, VTON~\cite{han2018viton} requires the textures, patterns, silhouettes, and design identity of a garment to be accurately preserved when it is transferred onto a reference person. Despite the strong generative capabilities of large-scale Diffusion Transformers (DiTs)~\cite{guo2025any2any,stylevton,kim2026coral,zhu2023tryondiffusion,kim2024stableviton}, achieving consistently high fidelity remains challenging. Existing models can introduce artifacts or degrade garment features, causing the generated garment to differ from the reference in color, texture, or fine detail. We argue that these limitations are partly architectural: many generative frameworks are designed for general-purpose editing and are not well suited to the strict identity-preservation requirements of VTON.

General-purpose models typically prioritize flexibility and global semantic alignment, whereas VTON requires fidelity to a specific reference garment. We argue that a key bottleneck in DiT-based VTON methods is the symmetric interaction pattern used in standard joint attention. In models such as FLUX.1 Kontext~\cite{labs2025fluxkontext}, noise and condition tokens are treated similarly and processed through bidirectional attention between the two streams. However, \textbf{VTON is an asymmetric information task}: the noise latent is stochastic and requires guidance from the condition, while the garment condition is deterministic and should remain largely unaffected by noise. To analyze this interaction, we introduce two information-theoretic indicators: (1) Conditional Attention Entropy (CAE), which measures feature unbiasedness by quantifying semantic dispersion and potential degradation in conditional features, and (2) Injected Information Flux (IIF), which measures injection effectiveness by quantifying the cumulative signal transferred from the condition to the noise latent.

The empirical results in Figure~\ref{fig: intro} suggest that standard bidirectional DiTs face an information bottleneck characterized by high CAE and low IIF. This bottleneck involves two attention paths. (1) The Condition-on-Noise (C-on-N) path allows deterministic garment tokens to attend to the stochastic noise latent. This reverse information flow increases CAE and corrupts the unbiased garment representation, causing conditional features to drift from the reference during denoising. (2) The Noise-on-Condition (N-on-C) path transfers garment information from the condition to the noise latent and directly determines IIF. In practice, timestep-dependent modulation designed for noisy latents can attenuate the conditional signal during critical stages of denoising. The resulting low IIF provides insufficient guidance for faithful reconstruction of the target garment.

Building on these insights, we propose RealFit, a high-fidelity VTON framework that addresses this bottleneck through two complementary mechanisms. (1) Unidirectional Information Flow (UIF): We block the C-on-N path while allowing noise tokens to attend to the garment condition. The condition attends only to itself and remains isolated from the stochastic noise latent. This design minimizes CAE, preserves high-frequency details, and maintains an unbiased garment representation throughout denoising.
(2) Decoupled Timestep Modulation (DTM): We decouple the modulation of the noise and condition branches to maximize IIF. Rather than following a dynamic schedule, the condition branch uses a fixed, optimal modulation state that maintains a strong conditional signal. This allows deterministic garment features to guide the entire denoising trajectory, improving visual fidelity.

Together, UIF and DTM make the condition branch independent of both the denoising timestep and the noise latent, so its features remain time-invariant. We use this property to implement a conditional KV cache: garment features are computed once and reused throughout denoising, avoiding redundant computation in subsequent iterations. This design accelerates inference and reduces GPU memory usage. Our contributions are as follows:
\begin{itemize}
\item We formulate VTON as an asymmetric information task based on the distinction between stochastic noise latents and deterministic conditional signals. This perspective motivates a diagnostic framework based on CAE and IIF for quantifying feature unbiasedness and injection effectiveness in generative models.

\item We propose RealFit, which combines UIF and DTM to address the limitations of symmetric DiT architectures. By protecting the deterministic condition from stochastic corruption and maximizing injection effectiveness, RealFit overcomes these structural bottlenecks while preserving garment identity.

\item Quantitative benchmarks and extensive qualitative evaluations show that RealFit achieves state-of-the-art performance. Its conditional KV cache reduces inference time by approximately 75\%, making high-fidelity generation more practical for deployment.
\end{itemize}

\section{Related Work}

\subsection{General-Purpose Image Editing}

General-purpose image editing has advanced substantially with the introduction of Diffusion Transformers (DiTs)~\cite{peebles2023scalable}. DiT-based frameworks such as FLUX.1 Kontext~\cite{labs2025fluxkontext} and OmniGen~\cite{xiao2024omnigen} use symmetric joint attention for in-context editing and instruction following without auxiliary conditioning networks. UniVG~\cite{fu2025univg} and Lumina-Image 2.0~\cite{qin2025lumina} process multimodal signals as joint sequences, supporting diverse tasks with a single set of model weights. The Show-o series~\cite{xie2025showo, xie2025showo2} unifies autoregressive modeling and discrete diffusion to handle interleaved modalities. ICEdit~\cite{zhang2025incontextedit} improves instruction adherence through in-context generation with minimal parameter-efficient fine-tuning. Training-free methods such as Stable Flow~\cite{avrahami2025stableflow} and SpotEdit~\cite{qin2025spotedit} enable localized editing by identifying critical layers or skipping computation in unchanged regions.

However, the symmetric bidirectional interactions used by these architectures are poorly suited to VTON. The garment condition must remain deterministic and retain its details, yet standard DiTs allow stochastic noise to corrupt conditional features and blur fine textures. Building on FLUX.1 Kontext, we introduce unidirectional information flow and decoupled modulation to preserve conditional integrity and maintain an unbiased garment representation.

\subsection{Virtual Try-On}
Diffusion models have become a dominant approach to generative modeling, with strong performance in high-fidelity image synthesis~\cite{rombach2022ldm, blattmann2023videoldm}, structured image editing~\cite{hertz2022p2p, meng2021sdedit}, and spatially conditioned generation~\cite{zhang2023controlnet}. Image-based VTON aims to generate an image of a person wearing a target garment while preserving both the person's identity and garment fidelity. Early methods typically relied on explicit geometric warping and extensive preprocessing. More recent methods build on large pretrained diffusion models and incorporate garment and person conditions through attention, improving photorealism and robustness. LADI-VTON~\cite{morelli2023ladi} uses textual inversion for guidance, while OOTDiffusion~\cite{xu2024ootdiffusion} and IDM-VTON~\cite{choi2024idm} use pretrained latent diffusion backbones to align and inject garment details through outfitting fusion or auxiliary modules, avoiding explicit warping. CatVTON~\cite{chong2025catvton} simplifies the architecture by spatially concatenating inputs within adapted self-attention modules, achieving competitive efficiency without redundant encoders. ITA-MDT~\cite{hong2025itamdt} uses image-timestep adaptive feature aggregation to capture fine details in salient garment regions. Other methods emphasize scalability and versatility: AnyFit~\cite{li2024anyfit} introduces parallel attention blocks for compositional clothing, and Any2AnyTryon~\cite{guo2025any2any} reduces reliance on auxiliary signals through adaptive positional embeddings. StyleVTON~\cite{stylevton} supports cross-domain story visualization, while FitDiT~\cite{jiang2024fitdit} allocates more capacity to high-resolution features using frequency-domain objectives. CORAL~\cite{kim2026coral} improves person-garment correspondence in unpaired settings through attention entropy minimization.

RealFit uses attention entropy to characterize the unbiasedness of deterministic conditional features. Our focus is on protecting reference features from stochastic interference and preserving their information throughout denoising.

\section{Methodology}

\subsection{Problem Formulation}
RealFit performs high-fidelity VTON using a garment image $x_g$ and a reference person image $x_r$. We generate a try-on image $x_p$ in which the person wears the garment in $x_g$, while preserving the person's identity, pose, body shape, and surrounding scene.

RealFit is built on a FLUX-style DiT architecture~\cite{labs2024flux}. A standard FLUX-style DiT processes the noise latent $\mathbf{x}_t$ and the visual condition $\mathbf{c}$ derived from $x_g$ and $x_r$ symmetrically, using bidirectional attention and shared timestep-dependent modulation. FLUX contains both double-stream and single-stream blocks, which differ primarily in how the visual and text modalities interact. Our design targets visual features in both block types.

\begin{figure*}[t]
  \centering
\includegraphics[width=1.0\textwidth]{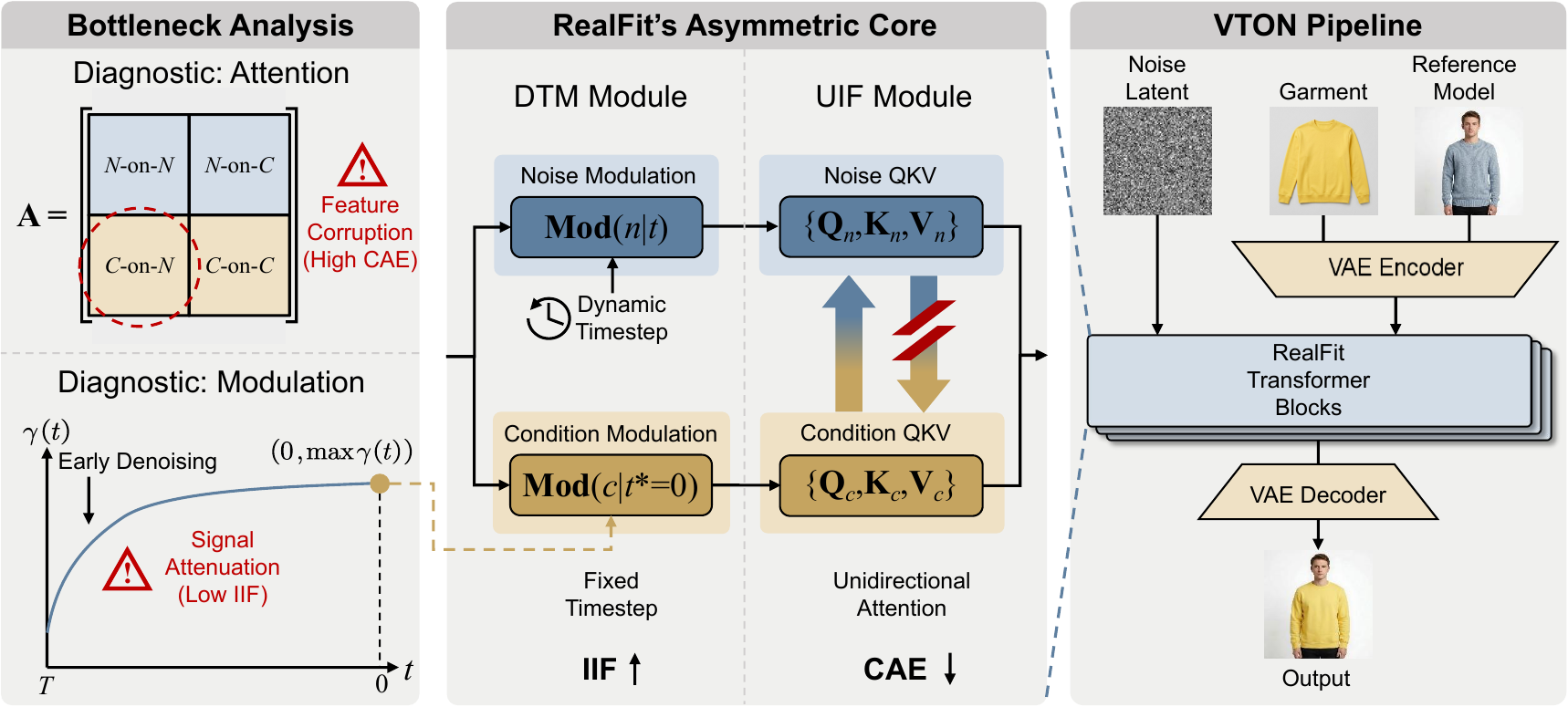}
  \caption{Overview of RealFit. High CAE in attention and low IIF in modulation can create an information bottleneck. RealFit addresses this bottleneck through an asymmetric design that combines UIF and DTM for high-fidelity VTON.}
  \label{fig: method}
\end{figure*}

\begin{figure*}[t]
\centering
\includegraphics[width=\textwidth]{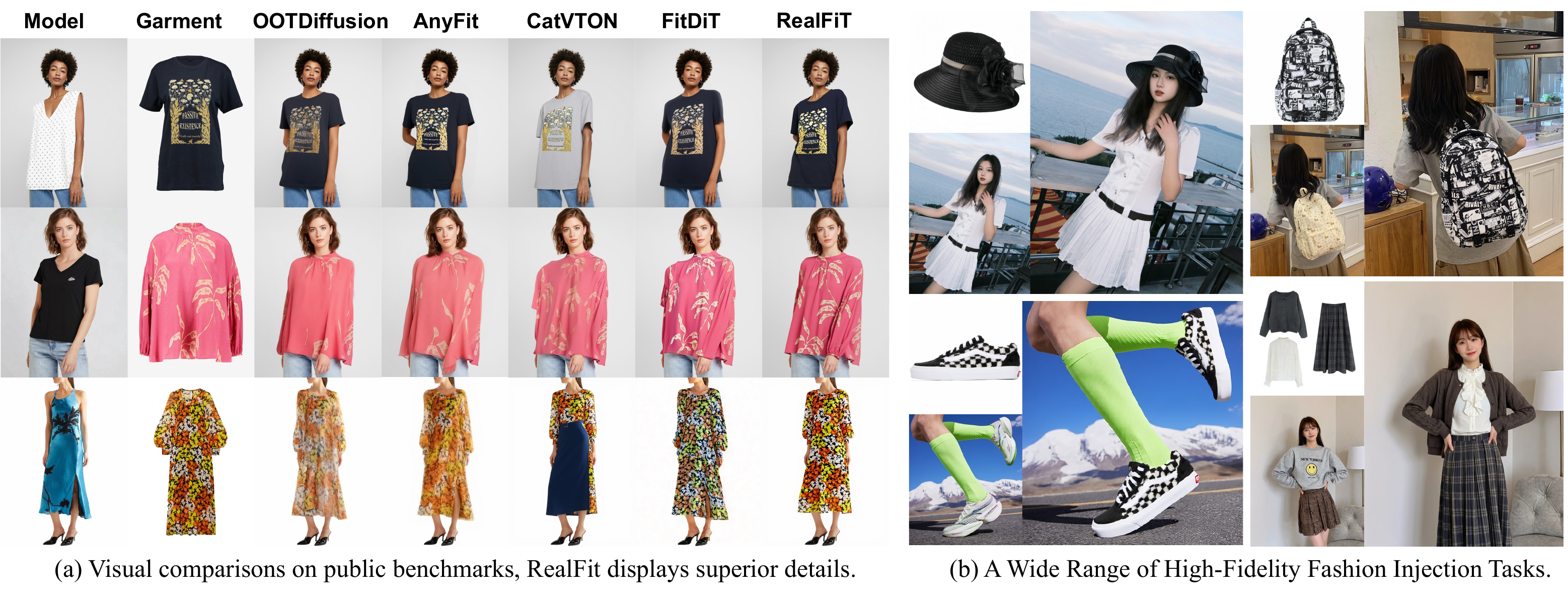}
\caption{Qualitative results. (a) RealFit protects deterministic garment features, achieving higher fidelity and sharper textures than the baselines. (b) Generalization across fashion items demonstrates broad applicability.}
\label{fig:Qualitative}
\end{figure*}

\begin{table*}[t]
\centering
\caption{Results on VITON-HD and DressCode. Best and second-best results are in \textbf{bold} and \underline{underlined}, respectively.}
\resizebox{\textwidth}{!}{
\begin{tabular}{lcccccccccccccccc}
\toprule
\multirow{3}{*}{Method} & \multicolumn{6}{c}{VITON-HD}  & \multicolumn{6}{c}{DressCode} \\
\cmidrule(lr){2-7} \cmidrule(lr){8-13}
& \multicolumn{4}{c}{Paired} & \multicolumn{2}{c}{Unpaired} & \multicolumn{4}{c}{Paired} & \multicolumn{2}{c}{Unpaired} \\
\cmidrule(lr){2-5} \cmidrule(lr){6-7} \cmidrule(lr){8-11} \cmidrule(lr){12-13}
& SSIM $\uparrow$ & LPIPS $\downarrow$ & FID $\downarrow$ & KID $\downarrow$ & FID $\downarrow$ & KID $\downarrow$ & SSIM $\uparrow$ & LPIPS $\downarrow$ & FID $\downarrow$ & KID $\downarrow$ & FID $\downarrow$ & KID $\downarrow$ \\
\midrule
LADI-VTON           & 0.875           & 0.091              & 9.42             & 1.63             & 14.64            & 8.75             & 0.902           & 0.052              & 6.94             & 2.33             & 10.67            & 5.78             \\
IDM-VTON            & 0.881           & 0.078              & 9.12             & 1.03             & 9.84             & 1.12             & 0.923           & 0.046              & 5.32             & 1.24             & 9.54             & 4.32             \\
AnyFit              & 0.893           & 0.075              & 8.60             & 0.55             & \underline{8.06} & 1.37             & 0.904           & 0.044              & 4.51             & \underline{0.48} & 7.33             & 2.59             \\
OOTDiffusion        & 0.878           & 0.071              & 8.81             & 0.82             & 12.40            & 4.68             & 0.885           & 0.053              & 4.61             & 0.95             & 12.56            & 6.62             \\
CatVTON            & 0.870           & \underline{0.056}  & 5.42             & 0.41             & 9.01             & 1.09             & 0.892           & 0.045              & 3.99             & 0.81             & 6.13             & 1.40             \\
ITA-MDT            & 0.885           & 0.083              & 5.46             & 0.63             & 8.67             & 1.21             & 0.922           & 0.055              & 6.16             & 0.94             & 11.02            & 1.75             \\
FitDiT             & \underline{0.898} & 0.066            & \underline{4.73} & \underline{0.19}    & 8.20             & \underline{0.34}    & \underline{0.925} & \underline{0.043} & \underline{2.63}   & 0.49             & \underline{4.73} & \underline{0.90} \\
\midrule
\textbf{RealFit (Ours)}                & \textbf{0.907}  & \textbf{0.049}     & \textbf{4.29}    & \textbf{0.15} & \textbf{7.74}   & \textbf{0.25} & \textbf{0.933}  & \textbf{0.034}     & \textbf{2.31} & \textbf{0.41}    & \textbf{3.96}    & \textbf{0.75}    \\
\bottomrule
\end{tabular}
}
\label{tab:main_results}
\end{table*}

\subsection{Bottlenecks in Symmetric Interaction}
\label{method32}

Standard DiT-based generation and editing models use symmetric interactions between noise and condition tokens. Treating these tokens equivalently supports flexible editing but conflicts with the strict fidelity requirements of VTON. The two streams represent distinct information states: the noise latent $\mathbf{x}_t$ is stochastic and requires structural guidance, whereas the visual condition $\mathbf{c}$ is a deterministic reference that encodes pixel-level identity. Symmetric attention allows noise to influence the condition, forcing conditional features to adapt to intermediate noisy states and potentially corrupting them. We formalize this interaction as follows.

Let $\mathbf{N} \in \mathbb{R}^{L_n \times d}$ and $\mathbf{C} \in \mathbb{R}^{L_c \times d}$ denote the noise and condition token sequences, respectively. The symmetric joint-attention matrix $\mathbf{A}$ is given by
\begin{equation}
\mathbf{A} = \text{Softmax}\left(\frac{\mathbf{Q}\mathbf{K}^\top}{\sqrt{d}}\right) =
\begin{bmatrix}
\mathbf{A}_{N \text{-on-} N} & \mathbf{A}_{N \text{-on-} C} \\
\mathbf{A}_{C \text{-on-} N} & \mathbf{A}_{C \text{-on-} C}
\end{bmatrix},
\end{equation}
where $\mathbf{A}_{I \text{-on-} J}$ denotes the attention weights between queries from $I$ and keys from $J$. This decomposition reveals two aspects of the information bottleneck.

\textbf{Conditional Attention Entropy}\hspace{0.8em}The $\mathbf{A}_{C \text{-on-} N}$ path allows stochastic noise to corrupt the deterministic condition. For a condition query $i$, the updated feature is a weighted sum of condition and noise values:
\begin{equation}
\mathbf{c}'_i = \sum_{j \in \text{Condition}} \mathbf{A}_{i,j} \mathbf{v}_j + \sum_{k \in \text{Noise}} \mathbf{A}_{i,k} \mathbf{v}_k,
\label{eq:update_rule}
\end{equation}
where the second term introduces variability from the stochastic latent. For high-fidelity VTON, we argue that the condition stream should remain unbiased: each condition token should concentrate its attention on a sparse set of deterministic keys to preserve high-frequency details.

We define Conditional Attention Entropy (CAE) as the sum of Shannon entropies over all condition-query attention distributions. CAE quantifies how concentrated these distributions are and characterizes the preservation of deterministic conditional features during joint attention:
\begin{equation}
\text{CAE} = \sum_{i=L_n+1}^{L_n+L_c} \left( -\sum_{j=1}^{L_n+L_c} \mathbf{A}_{i,j} \ln \mathbf{A}_{i,j} \right).
\end{equation}

Higher entropy indicates a more diffuse distribution. In Equation~\ref{eq:update_rule}, high CAE indicates dispersed attention, which can dilute garment identity when condition queries attend to stochastic noise rather than deterministic garment features. Decoupling the information pathways and blocking the C-on-N path prevents this reverse influence. Minimizing CAE supports conditional integrity and preserves detailed, unbiased garment features throughout denoising.

\textbf{Injected Information Flux}\hspace{0.8em}In addition to protecting the condition, high-fidelity VTON requires noise tokens to receive the garment's deterministic details effectively. For a noise query $i \in \{1, \dots, L_n\}$, a symmetric layer updates its feature $\mathbf{n}'_i$ as follows:
\begin{equation}
\mathbf{n}'_i = \underbrace{\sum_{j=1}^{L_n} \mathbf{A}_{i,j} \mathbf{v}_j}_{\text{Self-refining}} + \underbrace{\sum_{k=L_n+1}^{L_n+L_c} \mathbf{A}_{i,k} \mathbf{v}_k}_{\text{Injected Signal}},
\label{eq:noise_update}
\end{equation}
where the second term injects garment information into the noise stream. To quantify this transfer, we define Injected Information Flux (IIF) as the sum of attention weights $\mathbf{A}_{i,k}$ over the localized garment region. IIF measures how effectively the conditional signal guides the noise latent during denoising.

In standard denoising DiTs, query and key magnitudes are modulated by a shared timestep-dependent scale $\gamma(t)$, for example through AdaLN. Because $\mathbf{A}_{i,k} \propto \exp(\mathbf{q}_i^\top \mathbf{k}_k / \sqrt{d})$, IIF is sensitive to the norm of the condition keys $\|\mathbf{k}_k\|$, which is scaled by $\gamma(t)$. Under common modulation schedules, $\gamma(t)$ is relatively small early in denoising (large $t$) and increases as $t \to 0$. This time-dependent suppression may be unnecessary for a deterministic garment condition, whose information content remains relatively stable. It can weaken conditional guidance precisely when the latent is establishing the garment's global structure, potentially causing structural hallucinations or identity drift. We therefore decouple condition modulation from noise modulation and fix the condition branch to the maximal-scale state ($t=0$). This encourages consistently stronger IIF and helps preserve structural integrity from the earliest denoising steps.

\subsection{Asymmetric Information for VTON}
RealFit (Figure~\ref{fig: method}) addresses these bottlenecks with unidirectional attention and decoupled modulation.

\textbf{Unidirectional Information Flow}\hspace{0.8em}To reduce CAE and preserve feature unbiasedness, we replace symmetric interactions with unidirectional attention. A structured attention mask retains the N-on-N, N-on-C, and C-on-C paths while blocking the C-on-N path. This confines condition-token interactions to the deterministic condition stream, shielding it from stochastic noise. Unidirectional Information Flow (UIF) thus helps preserve the reference garment's identity throughout denoising.

\textbf{Decoupled Timestep Modulation}\hspace{0.8em}To increase IIF and strengthen conditional injection, we modulate the noise and condition branches separately. The noise branch follows the current timestep $t$ to accommodate changing noise levels. The condition branch uses a fixed timestep $t^\star$, which yields a relatively large, stable modulation scale throughout denoising. Keeping noise modulation dynamic and condition modulation fixed helps maintain the magnitude of condition keys and limits signal attenuation. DTM thus strengthens N-on-C attention and increases IIF to preserve fine garment details.

\textbf{Asymmetric Text Coupling}\hspace{0.8em}To maintain both instruction adherence and visual fidelity, we extend the asymmetric design to the text stream. The noise branch interacts with a text representation conditioned on the current timestep to support semantic planning throughout denoising. The condition branch instead uses a text representation conditioned on a fixed timestep, providing stable identity constraints from the garment description. This improves cross-modal consistency, allowing text and visual conditions to jointly guide generation.

\textbf{Time-Invariant KV Cache}\hspace{0.8em}The asymmetric design makes the condition branch independent of both the noise latent and the denoising timestep $t$, so its features remain constant during inference. We therefore compute the condition keys ($\mathbf{K}_c$) and values ($\mathbf{V}_c$) once at the initial step and cache them for subsequent iterations. This conditional KV cache eliminates redundant computation and reduces end-to-end inference time by over 75\% while preserving generation quality.

\section{Experiments}
\subsection{Experimental Settings}

\textbf{Datasets} We evaluate RealFit on two widely used public benchmarks, VITON-HD~\cite{choi2021vitonhd} and DressCode~\cite{morelli2022dresscode}, as well as an additional fashion dataset collected to evaluate generalization across diverse fashion items. VITON-HD contains 11,647 training and 2,032 test samples of women's upper-body garments. DressCode contains 48,392 training and 5,400 test samples spanning upper-body garments, lower-body garments, and dresses. To address the lack of reference person images in these benchmarks, we use open-source editing models to augment the original garment--ground-truth pairs into garment--reference person--ground-truth triplets. The reference and ground-truth images depict the same person in different clothing. We will release the augmented dataset to support further VTON research.

\noindent
\textbf{Implementation Details} We train RealFit for 40 epochs using the AdamW optimizer~\cite{kingma2014adam} with a learning rate of $3\times10^{-5}$. Training uses DeepSpeed ZeRO-2~\cite{rasley2020deepspeed, rajbhandari2020zero} on 8 NVIDIA H200 GPUs with a total batch size of 48. At inference time, we use 25 sampling steps on a single NVIDIA RTX 4090 GPU.

\noindent
\textbf{Evaluation Metrics}
Following prior work, we evaluate generation quality in both paired and unpaired settings. In the paired setting, we use SSIM~\cite{wang2004image}, LPIPS~\cite{zhang2018unreasonable}, FID~\cite{heusel2017gans}, and KID~\cite{binkowski2018demystifying}. In the unpaired setting, we use FID and KID to assess overall fidelity and diversity. To evaluate garment-level perceptual fidelity, we use Gemini 3.1 Pro Preview to score four attributes: detail, texture, silhouette, and color. We also report sample-averaged CAE and IIF in selected experiments, unless stated otherwise.

\subsection{Qualitative Results}

We qualitatively compare RealFit with representative methods, focusing on garment fidelity and robustness across try-on scenarios.

Figure~\ref{fig:Qualitative}(a) compares RealFit with representative baselines, including OOTDiffusion~\cite{xu2024ootdiffusion}, AnyFit~\cite{li2024anyfit}, CatVTON~\cite{chong2025catvton}, and FitDiT~\cite{jiang2024fitdit}. Existing symmetric methods often preserve the overall garment category but blur textures or alter garment identity when handling complex patterns and fine details. These artifacts are primarily associated with information corruption and signal attenuation in symmetric interactions. RealFit produces sharper textures and preserves garment identity more faithfully. UIF protects deterministic garment features, while DTM strengthens information injection, allowing the model to retain reference details and naturally adapt the garment to the person's body and pose.

To assess the versatility of the asymmetric framework, we also evaluate RealFit on challenging fashion categories using collected in-the-wild images. As shown in Figure~\ref{fig:Qualitative}(b), RealFit produces consistent, realistic results for diverse items, including hats, bags, shoes, and full suits. It preserves their distinctive material textures and fine structural details. These results support our central premise: preserving the condition as a deterministic signal and injecting its information effectively are important for high-fidelity conditional generation. They highlight RealFit's potential in virtual fashion.

\subsection{Quantitative Results}
Table~\ref{tab:main_results} compares RealFit with the baselines on VITON-HD and DressCode. RealFit outperforms the compared methods on all reported metrics, achieving state-of-the-art results.

RealFit achieves the highest SSIM scores, indicating better preservation of garment structure. Its lower LPIPS scores also reflect improved perceptual similarity and finer texture preservation. These results support our objective of high-fidelity information injection. RealFit additionally obtains the lowest FID and KID scores, indicating improved photorealism and closer alignment between the generated and real image distributions. This suggests that the model produces plausible try-on images while maintaining diversity. The gains in the unpaired setting further demonstrate robust generalization.

\begin{table}[t]
\centering
\caption{Ablation study of RealFit's components.}
\begin{adjustbox}{max width=\linewidth}
\begin{tabular}{c@{\,}lcccc}
\toprule
\multicolumn{2}{l}{\multirow{2}{*}{Method}} & \multicolumn{2}{c}{Generation} & \multicolumn{2}{c}{Information} \\
\cmidrule(lr){3-4} \cmidrule(lr){5-6}
\multicolumn{2}{c}{}                        & SSIM $\uparrow$ & LPIPS $\downarrow$ & CAE $\downarrow$ & IIF $\uparrow$ \\
\midrule
\multicolumn{2}{l}{Baseline}                & 0.854           & 0.156              & 8.0               & 438.7           \\
+ & Decouple                                  & 0.857           & 0.150              & 7.9               & 442.6           \\
+ & UIF                                       & 0.867           & 0.068              & 3.9               & 440.1           \\
+ & DTM                                       & 0.907  & 0.049     & 3.6      & 696.5  \\
\bottomrule
\end{tabular}
\end{adjustbox}

\label{tab:ab_modular}
\end{table}

\begin{table}[t]
\centering
\caption{Effects of UIF and DTM across backbones.}
\resizebox{\linewidth}{!}{
\small
\begin{tabular}{llcccc}
\toprule
Backbone & Config & SSIM$\uparrow$ & LPIPS$\downarrow$ & FID$\downarrow$ & KID$\downarrow$ \\
\midrule
\multirow{2}{*}{SDXL} & Baseline & 0.826 & 0.491 & 7.75 & 2.33 \\
 & + UIF \& DTM & \textbf{0.896} & \textbf{0.078} & \textbf{4.87} & \textbf{0.31} \\
\midrule
\multirow{2}{*}{SD3-DiT} & Baseline & 0.843 & 0.231 & 7.46 & 1.63 \\
 & + UIF \& DTM & \textbf{0.901} & \textbf{0.057} & \textbf{4.52} & \textbf{0.28} \\
\bottomrule
\end{tabular}}
\label{tab:various_backbones}
\end{table}

\begin{figure}[t]
  \centering
  \begin{minipage}{0.48\linewidth}
    \centering
    \includegraphics[width=\linewidth]{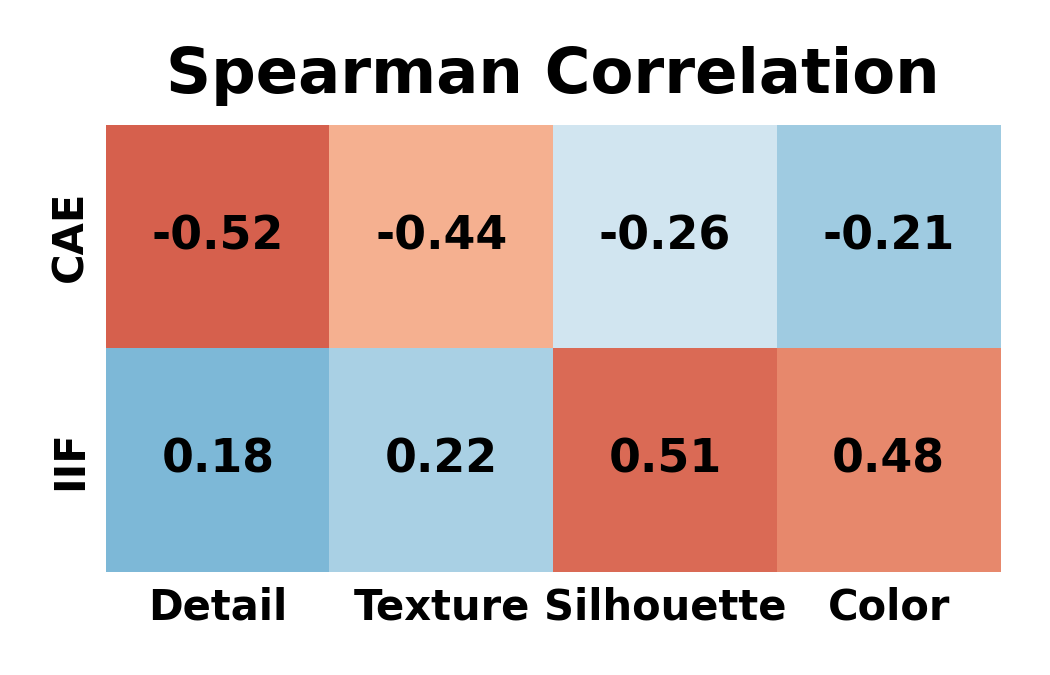}
    \caption{Spearman correlations between CAE/IIF and perceptual fidelity.}
    \label{fig:left}
  \end{minipage}
  \hfill
  \begin{minipage}{0.48\linewidth}
    \centering
    \includegraphics[width=\linewidth]{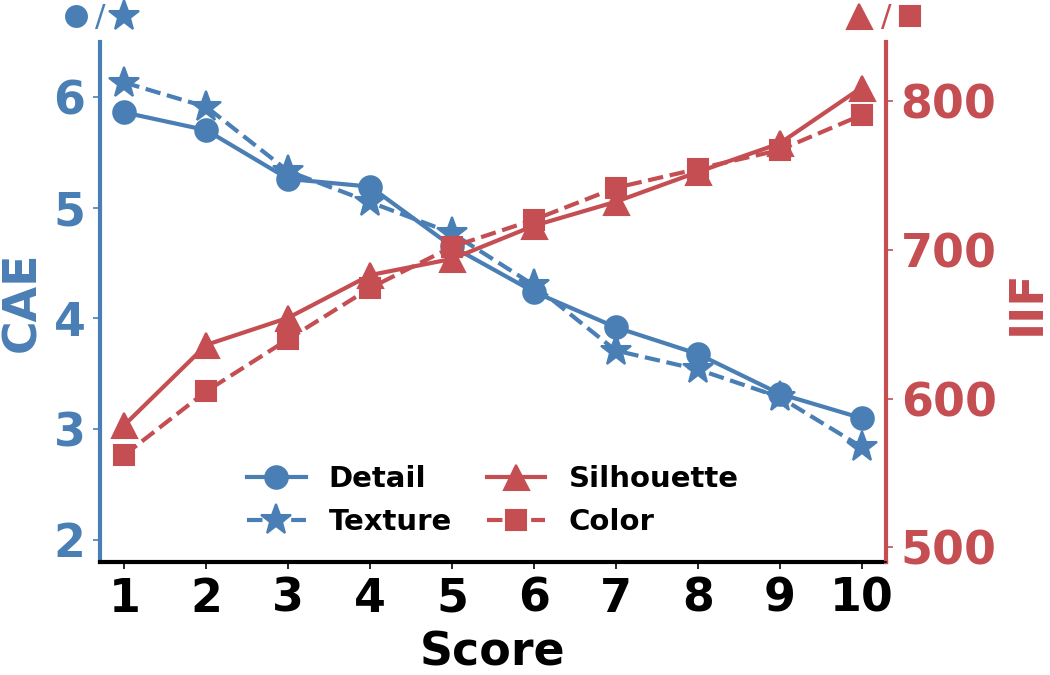}
    \caption{CAE/IIF trends with perceptual fidelity.}
    \label{fig:right}
  \end{minipage}
\end{figure}

\subsection{Ablation Studies}
We conduct ablation studies on VITON-HD to examine the contribution of each component. As shown in Table~\ref{tab:ab_modular}, the components provide incremental, complementary gains. Row 1 reports the standard symmetric DiT baseline. Row 2 decouples the QKV projections of the noise and condition branches, yielding a modest improvement. This may reflect the distributional difference between noise latents and garment features: separate projections allow the model to encode each stream more effectively. Adding UIF in Row 3 substantially reduces both LPIPS and CAE, indicating that unidirectional attention limits semantic corruption and feature dispersion while preserving conditional unbiasedness. Adding DTM in Row 4 further improves SSIM and IIF, demonstrating that branch-specific modulation strengthens the conditional signal and improves injection effectiveness.

Table~\ref{tab:various_backbones} evaluates the proposed design with different backbones. Applying UIF and DTM yields substantial improvements on both SDXL and SD3-DiT, supporting the effectiveness of asymmetric information injection across architectures.

\subsection{Perceptual Fidelity and CAE/IIF}

We further examine the relationship between CAE/IIF and perceptual fidelity. Using Gemini 3.1 Pro Preview~\cite{gemini_pro_3_preview}, we score 2,032 VITON-HD samples on detail, texture, silhouette, and color. Regression analyses relating these scores to CAE and IIF yield $p$-values below 0.05 in all cases, indicating statistically significant associations with generation quality. Figure~\ref{fig:left} reports the Spearman correlations between the two indicators and the four fidelity scores. CAE is more strongly correlated with high-frequency attributes, namely detail and texture, whereas IIF is more strongly correlated with low-frequency attributes, namely silhouette and color. This suggests that the indicators capture complementary aspects of perceptual fidelity, consistent with our analysis: optimizing CAE helps preserve a detailed, unbiased condition, while optimizing IIF provides the noise latent with sufficient deterministic guidance. Figure~\ref{fig:right} shows how each indicator varies with the fidelity scores most strongly associated with it. The trends support reducing CAE and increasing IIF as an effective optimization direction.

\subsection{Computational Efficiency}

\begin{table}[t]
\centering
\caption{Computational cost and inference time with and without the conditional KV cache.}
\begin{adjustbox}{max width=\linewidth}
\begin{tabular}{lccccc}
\toprule
Methods   & FLOPs (T) & Memory (MB) & Time (s) \\
\midrule
w/o Cache  & 0.052     & 25032          & 102.2         \\
w/ Cache     & \textbf{0.027}     & \textbf{19436}          & \textbf{24.7}        \\
\bottomrule
\end{tabular}
\end{adjustbox}
\label{tab:efficiency}
\end{table}

We evaluate RealFit's computational cost, memory usage, and inference time. Because the condition branch is independent of the noise latent and timestep $t$, its keys and values can be computed once and reused throughout denoising. As shown in Table~\ref{tab:efficiency}, the conditional KV cache removes redundant computation and reduces end-to-end inference time by approximately 75\% without compromising fidelity. It also reduces peak GPU memory usage by avoiding repeated processing of condition tokens, improving the practicality of deployment.

\section{Conclusion}

We present RealFit, a high-fidelity VTON framework based on asymmetric information dynamics in DiTs. Two diagnostic indicators, Conditional Attention Entropy and Injected Information Flux, characterize the information bottleneck in symmetric architectures. Unidirectional Information Flow protects deterministic garment features from stochastic noise, while Decoupled Timestep Modulation maintains a strong conditional signal. RealFit achieves state-of-the-art fidelity across benchmarks, and its time-invariant condition branch supports a conditional KV cache that reduces inference time by approximately 75\% without additional training. These results highlight the potential of asymmetric interactions for high-fidelity, identity-preserving generation.

\bibliographystyle{plainnat}
\bibliography{main}

\end{document}

%% file: authors.tex
\author[1]{Zishu Qin}
\author[1]{Zhiyu Jin}
\author[1]{Pipei Huang}
\author[1,\boxtimes]{Hao Zhou}

\affiliation[1]{Alibaba Group}

\contribution[\boxtimes]{Corresponding author}